\documentclass[10pt,twocolumn,letterpaper]{article}

\usepackage[pagenumbers]{cvpr} % To force page numbers, e.g. for an arXiv version

\definecolor{cvprblue}{rgb}{0.21,0.49,0.74}
\usepackage[pagebackref,breaklinks,colorlinks,allcolors=cvprblue]{hyperref}
\usepackage{bbding}
\usepackage[]{xcolor}  

\def\paperID{xxxx} % *** Enter the Paper ID here
\def\confName{CVPR}
\def\confYear{2025}

\definecolor{Gray}{gray}{0.9}

\def\eg{\emph{e.g}\onedot} 
\def\ie{\emph{i.e}\onedot}

\newcommand{\tablestyle}[2]{\setlength{\tabcolsep}{#1}\renewcommand{\arraystretch}{#2}\centering\footnotesize}
\usepackage{multirow}

\newcommand{\modelname}{CoCal\xspace}

\usepackage{colortbl}

\title{Dictionary-based Framework for Interpretable and Consistent Object Parsing}

\author{
    Tiezheng Zhang \quad Qihang Yu \quad Alan Yuille \quad Ju He \\
    Johns Hopkins University\\
    \url{https://github.com/ollie-ztz/CoCal}
}

\begin{document}
\maketitle
\begin{abstract}
In this work, we present \modelname, an interpretable and consistent object parsing framework based on dictionary-based mask transformer. Designed around \textbf{Co}ntrastive \textbf{C}omponents and Logic\textbf{al} Constraints, \modelname rethinks existing cluster-based mask transformer architectures used in segmentation; Specifically, \modelname utilizes a set of dictionary components, with each component being explicitly linked to a specific semantic class. To advance this concept, \modelname introduces a hierarchical formulation of dictionary components that aligns with the semantic hierarchy. This is achieved through the integration of both within-level contrastive components and cross-level logical constraints. Concretely, \modelname employs a component-wise contrastive algorithm at each semantic level, enabling the contrasting of dictionary components within the same class against those from different classes. Furthermore, \modelname addresses logical concerns by ensuring that the dictionary component representing a particular part is closer to its corresponding object component than to those of other objects through a cross-level contrastive learning objective. To further enhance our logical relation modeling, we implement a post-processing function inspired by the principle that a pixel assigned to a part should also be assigned to its corresponding object. With these innovations, \modelname establishes a new state-of-the-art performance on both PartImageNet and Pascal-Part-108, outperforming previous methods by a significant margin of 2.08\% and 0.70\% in part mIoU, respectively. Moreover, \modelname exhibits notable enhancements in object-level metrics across these benchmarks, highlighting its capacity to not only refine parsing at a finer level but also elevate the overall quality of object segmentation.

\end{abstract}    
\section{Introduction}
\label{sec:intro}

\begin{figure}[!ht]
    \centering
    \includegraphics[width=0.5\textwidth]{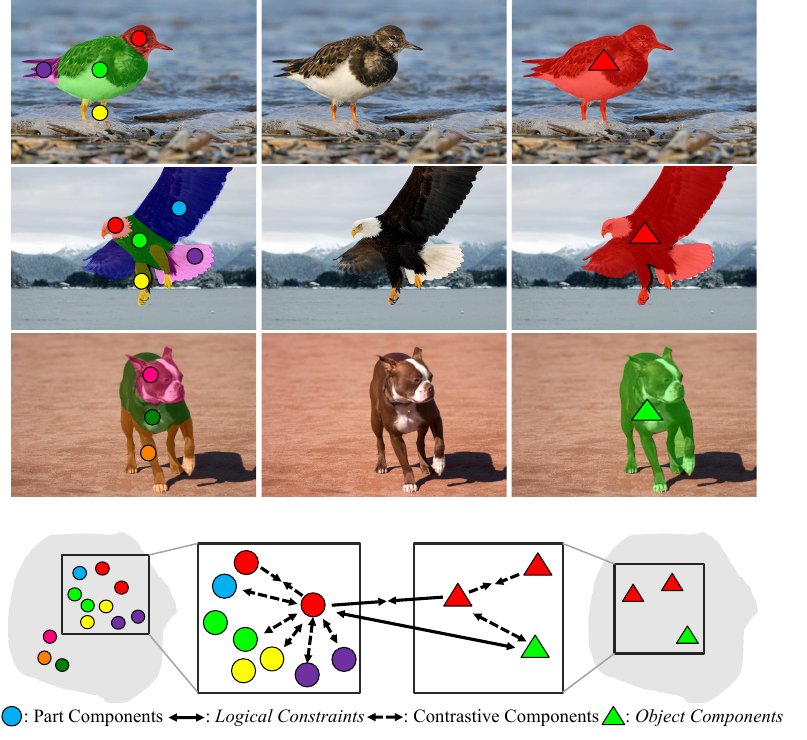}
    \caption{\textbf{Illustration of the proposed component-wise contrastive objectives}. \modelname establishes two discriminative dictionaries at the part and object levels. Within the same semantic level, part/object components of the same classes are pulled closer ($\rightarrow\leftarrow$), while those of different classes are pushed apart ($\leftarrow\rightarrow$) (\ie, contrastive components). At the cross-semantic level, part components and their corresponding object components are pulled closer and vice versa (\ie logical constraints).
    }
    \label{fig:teaser}
\end{figure}

Human perception involves the ability to decompose an object into its semantically meaningful components (\ie, parts). For instance, when observing a dog, humans not only identify it as a dog but also simultaneously discover its head, torso, and other components, facilitating a more interpretable and resilient understanding of real-world scenarios. More specifically, humans can estimate the pose of a dog by considering the spatial arrangement of its parts, even in instances where some parts may be missing. This comprehensive perception enables individuals to make judgments about the potential actions of the observed object.

By contrast, emulating this innate human visual capability presents a big challenge for modern computer vision models. The predominant focus within the field has been on addressing semantic segmentation at the object level, with minimal attention given to intermediate part representations. Notable works~\cite{de2021part,michieli2020gmnet,li2022panoptic,michieli2022edge} in object parsing primarily extend algorithms designed for general segmentation, overlooking the fact that parts, being at a lower semantic level, can be captured more efficiently and interpretably through clustering. As a result, these works often adhere to frameworks tailored for object segmentation without incorporating specialized designs for handling parts. Moreover, even though certain studies~\cite{eslami2012generative,wang2015joint,he2023compositor} highlight the mutual benefit between object parsing and object segmentation, they typically treat these semantic levels separately, disregarding the logical relationship between them. Consequently, the optimization objectives for these two levels are disjoint, leading to sub-optimal predictions.

In this work, we propose \modelname, a general dictionary-based framework aimed at addressing these challenges. \modelname is built on top of an off-the-shelf cluster-based mask transformer, utilizing a set of dictionary components where each component is explicitly associated with a specific semantic class to facilitate the grouping of pixels belonging to that class. This enables \modelname to conduct inference in a straightforward parameter-free manner through nearest neighbor search on the pixel feature maps within the class dictionary. Taking this concept further, \modelname introduces a hierarchical formulation of dictionary components, aligning with the semantic hierarchy, which naturally forms the logical paths within the structure (\eg, bird-head $\rightarrow$ bird).
\modelname advances the learning of the above formulation through two simple yet effective targets: learning contrastive objectives for obtaining discriminative dictionary components and exploring logical relations for consistent predictions. 
Specifically, as depicted in Fig.~\ref{fig:teaser}, at each semantic level, \modelname employs a component-wise contrastive algorithm to pull closer the dictionary components withing the same class while pushing away those from different classes, thus a better-structured dictionary space is derived, ultimately improving the performance of object parsing. Then to model the cross-level logical relations, \modelname further contrasts the positive pair between dictionary component representing a particular part and its corresponding object dictionary components against the negative pairs involving the part component and all other object components. 
For further enhancement of logical constraints during testing, \modelname implements a post-processing function inspired by the principle that a pixel of a given part class must also be predicted as its corresponding object class. More precisely, \modelname enables this ability by calculating the logical path probability through multiplying the part-level similarity and object-level similarity. Subsequently, \modelname assigns each pixel with the class labels in the top-scoring path. This approach effectively captures the cross-level semantic information and corrects potential cross-level inconsistencies during inference. In summary, our contributions in this work include:
\begin{enumerate}
    \item We present \modelname, a versatile dictionary-based framework tailored for object parsing and can be integrated with various cluster-based mask transformers.
    \item We propose a component-wise contrastive learning method designed to enhance the learning of discriminative dictionary components and foster the development of a well-structured dictionary space.
    \item We introduce logical constraints for object parsing, leveraging inherent semantic hierarchy information to alleviate cross-level inconsistency.
    \item We validate the effectiveness of \modelname through extensive experiments on PartImageNet and Pascal-Part-108. The incorporation of the above modules notably improves performance on both the part and the object level.
\end{enumerate}

\section{Related Work}
\label{sec:related}

\subsection{Object Parsing}
The extensive literature on object parsing can be divided into single-object multi-part parsing~\cite{hariharan2015hypercolumns,xia2016zoom,chen2016attention,liang2018look,xia2017joint,nie2018mutual} and multi-object multi-part parsing~\cite{zhao2019multi,michieli2020gmnet,singh2022float,he2023compositor}. Single-object multi-part parsing has primarily focused on specific classes, such as humans~\cite{zhu2011max,liang2015deep,yamaguchi2012parsing}, animals~\cite{wang2015semantic}, and vehicles~\cite{eslami2012generative,lu2014parsing,song2017embedding}. While the methodologies addressing multi-object multi-part parsing mainly focus on employing top-down or coarse-to-fine strategies. 
Specifically, Singh et al.~\cite{singh2022float} proposed FLOAT, a factorized top-down parsing framework by first detecting the object followed with zooming in for obtaining higher quality part masks. On the contrary, He et al.~\cite{he2023compositor} introduced Compositor, a bottom-up architecture designed to iteratively learn objects by clustering pixels to derive parts. 
Recently, there are also explorations in the closely related area of panoptic part segmentation within the research community. Notable works such as~\cite{de2021part,li2022panoptic,verboeket2022hierarchical,li2023panopticpartformer++,alt2023efficientpps,muralidhara2024jppf} have delved into the semantic parsing of objects while also distinguishing parts between different instances. However, a common trend in these works, whether focused on semantic object parsing or panoptic part segmentation, involves extending standard segmentation models, often overlooking the nuanced semantic levels of parts. In contrast, \modelname takes a novel approach by focusing specifically on semantic object parsing. It redefines the paradigm of cluster-based mask transformers and introduces a novel dictionary-based framework meticulously tailored for object parsing.

\subsection{Cluster-based Mask Transformer}
With the recent progress in transformers \cite{carion2020end}, a new paradigm named mask classification~\cite{wang2020axial,wang2021max,cheng2021per,strudel2021segmenter,zhang2021k,cheng2022masked} has been proposed, where segmentation predictions are represented by a set of binary masks with its class label, which is generated through the conversion of object queries to mask embedding vectors followed by multiplying with the image features. The predicted masks are trained by Hungarian matching with ground truth masks. Thus the essential component of mask transformers is the decoder which takes object queries as input and gradually transfers them into mask embedding vectors. Recently, cluster-based mask transformers are introduced in \cite{yu2022cmt,yu2022k,liang2023clustseg}, which rethinks the design of the decoder by replacing the cross-attention with a k-means~\cite{lloyd1982least} attention. Building upon these explorations, \modelname introduces a global class dictionary and replaces the Hungarian matching with a fixed one-to-one matching, thereby establishing an interpretable dictionary-based framework for part segmentation.

\subsection{Contrastive Learning in Segmentation}
Contrastive learning~\cite{robinson2020contrastive,chen2020simple,khosla2020supervised,chen2020big,he2020momentum,chen2020improved,he2022masked} has emerged as a prominent technique in computer vision as an effective method for learning feature representation for self-supervised models. The core idea lies in contrasting similar (positive) data pairs against dissimilar (negative) pairs. Recently, Wang et al.~\cite{wang2021exploring} raise a pixel-to-pixel contrastive learning method for semantic segmentation, which enforces pixel embeddings belonging to the same semantic class to be more similar than embeddings from different classes. \cite{du2022weakly,li2023contrast,chen2023pipa,lv2023confidence,wang2023space,xie2023sepico} are built upon this concept, extending it to various segmentation domains. Motivated by these advancements, we propose a component-wise contrastive learning method tailored for modern cluster-based mask transformers, which effectively learns discriminative dictionary components within the clustering scheme.  

\subsection{Logical Constraints in Segmentation}
Few segmentation models~\cite{liang2018dynamic,xiao2018unified,wang2019learning,li2020deep,wang2020hierarchical,li2022deep,ke2022unsupervised,li2023logicseg} consider the implicit logic rules inherent in structured labels. While the majority of them are dedicated to human parsing, a few recent works~\cite{li2022deep,li2023logicseg} tackle the general segmentation in a flexible function and avoid incorporating label taxonomies into the network topology. Concretely, Li et al.~\cite{li2022deep} enhance the logical consistency by modeling the segmentation as a pixel-wise multi-label classification. Li et al.~\cite{li2023logicseg} exploit neuro-symbolic computing for grounding logical formulae onto data. In contrast to these efforts, \modelname introduces an object level on top of the part and models logical rules as a contrastive objective during training. 

\section{Method}
\label{sec:method}

\begin{figure*}[t!]
    \centering
    \includegraphics[width=\textwidth]{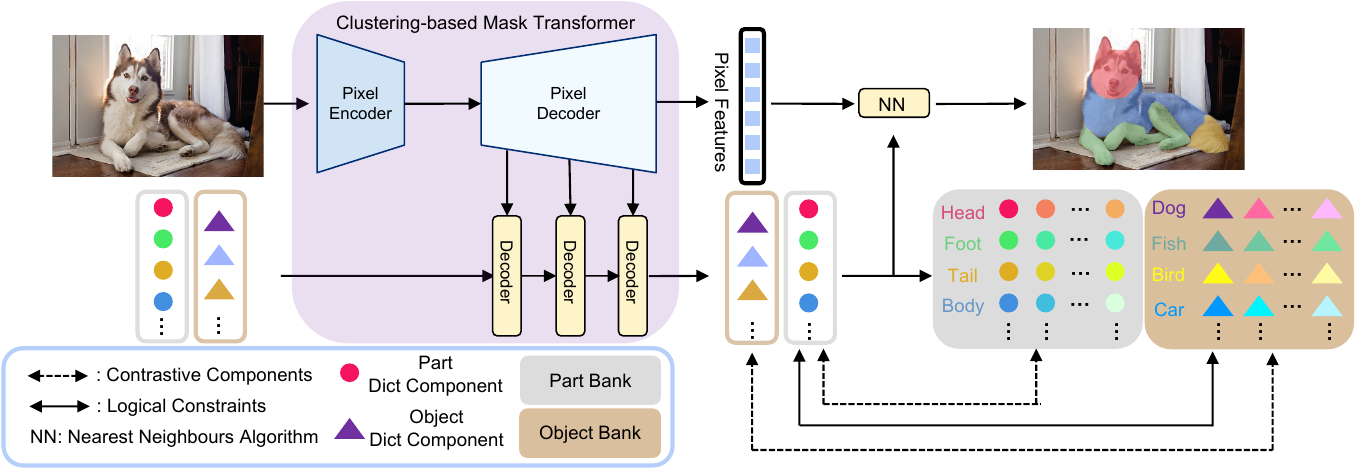}
    \caption{\textbf{Meta-architecture of the proposed \modelname}. \modelname builds on top of an off-the-shelf clustering-based mask transformer, incorporating dictionary components that function as the cluster centers for each semantic class. Throughout training, the dictionary components in \modelname are updated via both mask-wise objectives from the transformer and contrastive objectives from the dictionary. During testing, \modelname adopts a straightforward inference approach by executing nearest neighbor search of the pixel features on the dictionary components.}
    \label{fig:meta}
\end{figure*}

In this section, we begin with a brief overview of existing cluster-based mask transformer segmentation frameworks, providing context for the introduction of our key innovations. We then delve into the modifications we've made, particularly the integration of sets of dictionary components aligned with the semantic hierarchy. This tailored approach forms the basis of our dictionary-based mask transformer framework, specifically optimized for object parsing. Afterwards, our discussion focuses on two main aspects: the implementation of contrastive components, enhancing effectiveness and interpretability, and the incorporation of logical constraints, crucial for improving parsing consistency. Finally, we provide a detailed exploration of the meta-architecture of \modelname, elucidating the structural components and operational dynamics of the system.

\subsection{Recap of Cluster-based Mask Transformer}
\label{subsec:method-recap}
Cluster-based mask transformers~\cite{yu2022cmt,yu2022k,liang2023clustseg} have demonstrated considerable efficacy across a range of segmentation tasks. To provide a universal context, our discussion primarily focuses on semantic segmentation:

\noindent\textbf{Problem Statement}\quad
Semantic segmentation aims to divide an image $\mathbf{I} \in \mathbb{R}^{\mathit{H} \times \mathit{W} \times \mathit{3}}$ into distinct, non-overlapping masks, each associated with a semantic label. This process is formalized as follows:
\begin{equation}
\begin{aligned}
    \{y_{i}\}_{i=1}^{M_p} = \{(d_{i}, c_{i})\}_{i=1}^{M},
    \label{formula:mask_gt}
\end{aligned}
\end{equation}
where $d_{i} \in {\{0,1\}}^{H \times W}$ identifies whether a pixel is part of a specific region, $c_{i}$ represents the corresponding class label and $M$ denotes the total number of ground-truth masks.

In contrast to traditional approaches, cluster-based mask transformers generate a prediction set that mirrors the format of the ground-truth, comprising $N$ masks (where $N$ is a predetermined number satisfying $N \ge M$) along with their class associations:
\begin{equation}
\begin{aligned}
    \{\hat{y}_{i}\}_{pi=1}^{N} = \{(\hat{m}_{i}, \hat{c}_{i})\}_{pi=1}^{N}.
    \label{formula:mask_pred}
\end{aligned}
\end{equation}

These $N$ masks are derived from object queries that consolidate information from pixel features. The key distinction between cluster-based mask transformers and standard query-based transformers is evident in their respective updating mechanisms. Specifically, the query-based mask transformer updates the object queries as follows:

\begin{equation}
\begin{aligned}
    \mathbf{\hat{O}} &= \mathbf{O} + \operatornamewithlimits{softmax}_{HW}(\mathbf{Q}^{o} \times (\mathbf{K}^p)^\mathrm{T}) \times \mathbf{V}^p,
    \label{formula:cross-attn}
\end{aligned}
\end{equation}
while cluster-based mask transformer exploits:
\begin{equation}
\begin{aligned}
    \mathbf{\hat{O}} &= \mathbf{O} + \operatornamewithlimits{argmax}_{N}(\mathbf{Q}^{o} \times (\mathbf{K}^p)^\mathrm{T}) \times \mathbf{V}^p,
    \label{formula:kmeans-attn}
\end{aligned}
\end{equation}
where $\mathbf{O} \in \mathbb{R}^{N \times D}$ symbolizes the $N$ object queries with $D$ channels, and $\mathbf{\hat{O}}$ represents the updated queries. $\mathbf{Q}^{o} \in \mathbb{R}^{N \times D}, \mathbf{K}^p \in \mathbb{R}^{HW \times D}, \mathbf{V}^p \in \mathbb{R}^{HW \times D}$ represent the linearly projected features for the query, key, and value, respectively. The notations $HW$ and $N$ indicate the axes for the \textit{softmax} and \textit{argmax} operations on the pixel and query dimensions, respectively. The superscripts $p$ and $o$ denote the features projected from pixel features and object queries, correspondingly.

Intuitively, these update rules explicitly compute the affinity between object queries and pixel features (\ie, $\mathbf{Q}^{o} \times (\mathbf{K}^p)^\mathrm{T}$), followed by assigning a one-hot cluster assignment to each pixel via the \textit{argmax} operation. This assignment clusters affiliated pixel features to update the corresponding object queries. The updated queries $\mathbf{\hat{O}}$ are then used to generate the prediction set $\hat{y}$, which is matched with the ground-truth set $y$ through Hungarian Matching~\cite{Kuhn1955NAVAL} during training to compute the losses. For a more detailed exposition of cluster-based mask transformers, the reader is referred to kMaX-Deeplab~\cite{yu2022k}.

\subsection{Dictionary-based Mask Transformer Framework}
\label{subsec:method-dict}
Building upon the cluster-based mask transformers, we introduce the concept of dictionary-based mask transformer. This architecture primarily pivots on the integration of a set of dictionary components, which supersedes the use of object queries $C$ in traditional models. Specifically, the dictionary $\mathbf{C} \in \mathbb{R}^{P \times D}$ comprises $P$ learnable components, each dedicated to grouping pixels associated with a specific class, where $P$ also represents the number of classes.

A key distinction of the dictionary-based mask transformers, as compared to query-based or cluster-based mask transformers, lies in its structural efficiency. Traditional mask transformers typically encompass a larger number of object queries $\mathbf{O} \in \mathbb{R}^{N \times D}$ than the number of classes, necessitating the filtering of redundant queries through `void' classes. In contrast, our dictionary-based mask transformer maintains an exact one-to-one correspondence between the dictionary components and the classes. This direct alignment facilitates a streamlined training process, where $\mathbf{C}$ is updated following Eq.~\ref{formula:kmeans-attn}. Consequently, the Hungarian Matching process is replaced by fixed matching mechanism (\ie, $C_i$ corresponds to the cluster center of class $p_i$).
% , which is discussed in ~\ref{subsec:ablation-studies}.

In the testing phase, the dictionary-based mask transformer exhibits its efficiency through a parameter-free operation. It accomplishes this by conducting a per-pixel nearest neighbor search within the pixel feature maps, utilizing the dictionary $\mathbf{C}$. This method grants the dictionary-based mask transformer a cohesive, simplified, and easily interpretable architecture, both in training and testing, which is specially designed for object parsing.

\subsection{\modelname: Interpretable and Consistent Object Parsing}
\label{subsec:method-framework}
\noindent \textbf{Hierarchical Structure of Dictionaries Across Multiple Levels}\quad The classification labels for various parts inherently contain rich logical information within their structure. For example, the label `dog-head' logically suggests a closer relationship to `dog-torso' than to `fish-tail'. To utilize these implicit logical relationships inherent in structured labels, \modelname extends the dictionary-based mask transformer into a hierarchically structured framework.

Specifically, \modelname introduces an additional tier of object-level classes on top of the part-level classes, aligning with their semantic context. This structure mirrors the formulation used for parts, and we denote the object-level dictionary as $\mathbf{\widetilde{C}} \in \mathbb{R}^{\widetilde{P} \times D}$, where $\widetilde{P}$ is the number of learnable dictionary components corresponding to the number of object classes.

\noindent \textbf{Enhancing Dictionary Discrimination Through Contrastive Objectives}\quad
For the effective training of \modelname, we utilize contrastive learning to discern and learn discriminative dictionary components. The underlying principle is intuitive: components associated with the same class should exhibit similarity and, thus, are brought closer together, whereas those from different classes are separated.

Taking the part dictionary $\mathbf{C}$ as an example, \modelname incorporates a part memory bank $\mathbf{B} \in \mathbb{R}^{P \times S \times D}$, where $S$ represents the number of samples of each class stored in the dictionary. This memory bank stores the dictionary components for the observed parts during training. Given a ground-truth set $y$, \modelname retrieves the relevant dictionary components $\mathbf{C}(y)$ from $\mathbf{C}$. These components correspond to all the parts that have manifested in the training data. The contrastive loss is then computed based on these retrieved components.

This approach facilitates the creation of more distinct and separate clusters of dictionary components, thereby improving the accuracy and robustness of \modelname. By leveraging contrastive learning, \modelname not only distinguishes between different classes more effectively but also enhances the overall coherence and interpretability of the segmentation results. The contrastive loss is formulated as:
\begin{small}
\begin{equation}
    \mathcal{L}_{p\_con}(\mathbf{C}(y)) = \sum_{x \in M}\frac{-1}{|\mathbf{B}(x)|}\sum_{j \in \mathbf{B}(x)}\log\frac{\exp(\mathbf{C}(y)_i \cdot B_j / \tau)}{\sum_{k \in \mathbf{B}}\exp(\mathbf{C}(y)_i \cdot B_k / \tau)},
    \label{formula:contrast_p}
\end{equation}
\end{small}where $\mathbf{B}(x)$, $B_j$ and $\mathbf{C}(y)_i$ denote the memory bank components belonging to class $x$ in $\mathbf{B}$, memory bank components in $\mathbf{B}$ and dictionary components in $\mathbf{C}(y)$, respectively. $\tau \in R^+$ is a scalar temperature parameter. Motivated by~\cite{lin2017focal, robinson2020contrastive}, we additionally exploit hard negative mining to put more focus on hard examples, where we only select top-k hardest samples defined by the cosine similarity from $\mathbf{B}$ when calculating Eq. \ref{formula:contrast_p}. Similarly, we maintains the object memory bank $\widetilde{\mathbf{B}}$ and apply the same contrastive loss on the object dictionary $\mathbf{\widetilde{C}}$ as:
\begin{small}
\begin{equation}
    \mathcal{L}_{o\_con}(\mathbf{\widetilde{C}}(y)) = \sum_{x \in M}\frac{-1}{|\mathbf{\widetilde{B}}(x)|}\sum_{j \in \mathbf{\widetilde{B}}(x)}\log\frac{\exp(\mathbf{\widetilde{C}}(y)_i \cdot \widetilde{B}_j / \tau)}{\sum_{k \in \mathbf{\widetilde{B}}}\exp(\mathbf{\widetilde{C}}(y)_i \cdot \widetilde{B}_k / \tau)},
    \label{formula:contrast_o}
\end{equation}
\end{small}where $\mathbf{\widetilde{B}}(x)$, $\widetilde{B}_k$ and $\mathbf{\widetilde{C}}(y)_i$ denote the memory bank components belonging to class $x$ in $\mathbf{\widetilde{B}}$, memory bank components in $\mathbf{\widetilde{B}}$ and dictionary components in $\mathbf{\widetilde{C}}(y)$.

\begin{figure}[]
    \centering
    \includegraphics[width=0.5\textwidth]{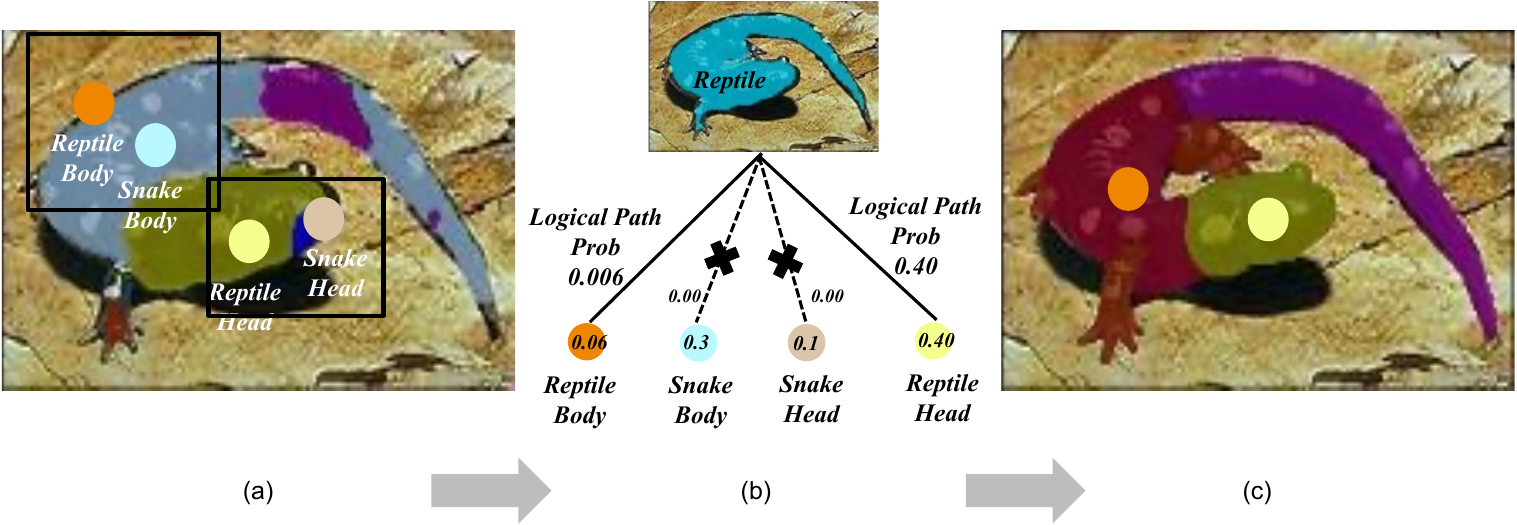}
    \caption{\textbf{Illustration of logical constraints at inference.} In this picture, a reptile-head and reptile-body are wrongly predicted as the snake-head and snake-body, respectively. \modelname corrects the wrong prediction by computing the logical path probability through multiplying the part-level probability and object-level probability and re-assigns the labels along the path thus producing the correct part prediction.}
    \label{fig:logical}
\end{figure}

\noindent \textbf{Logical constraints for consistent predictions}\quad
To alleviate the potential inconsistency in part class prediction within the same object or cross-level prediction, we explore logical constraints following the innate semantic hierarchy to encourage the consistency at training and put constraints at inference. Based on that, \modelname explores two crucial logical constraints. More specifically, motivated by the fact that the part dictionary components should be closer to its corresponding object dictionary components compared to other object dictionary components, we apply the cross-level contrastive loss as: 
\begin{small}
\begin{equation}
    \mathcal{L}_{logic}(\mathbf{C}(y)) = \sum_{x \in M}\frac{-1}{|\mathbf{\widetilde{B}}(x)|}\sum_{j \in \mathbf{\widetilde{B}}(x)}\log\frac{\exp(C(y)_i \cdot \widetilde{B}_j / \tau)}{\sum_{k \in \mathbf{\widetilde{B}}}\exp(C(y)_i \cdot \widetilde{B}_k / \tau)}.
    \label{formula:contrast_cross}
\end{equation}
\end{small}Note that Eq.~\ref{formula:contrast_cross} models the cross-level contrastive relations and encourages parts belonging to the same object to share similar features. As a result, different parts within one object will tend to have the same object class prediction thus effectively alleviates the inconsistency problem. Furthermore, \modelname takes the fact that if a pixel belongs to a certain part, it must also belongs to the corresponding object and models this as a post-processing function during testing. Concretely, as shown in Fig.~\ref{fig:logical}, \modelname first calculates the logical path probability through multiplying the part-level class probability and object-level class probability obtained through nearest neighbor search followed by assigning each pixel with the labels in the top-scoring path.

\noindent \textbf{Meta-Architecture Overview}\quad 
As illustrated in Fig.~\ref{fig:meta}, the meta-architecture of our proposed \modelname is a comprehensive framework that incorporates several crucial elements. It builds on top of an off-the-shelf cluster-based mask transformer, which is responsible for extracting pixel features. The core of the architecture is formed by the part and object dictionaries, crucial for storing discriminative dictionary components capable of grouping pixels based on their respective semantic classes. In tandem with these dictionaries, the part and object banks are meticulously designed to retain a history of observed components, a key element for contrastive loss calculation within and across semantic levels. Consequently, these modules collectively constitute \modelname, an innovative and cohesive dictionary-based framework for object parsing. This approach guarantees interpretability and consistency by embedding a logical, hierarchical structure into the segmentation process. Through this methodology, \modelname represents a significant advancement in object parsing, offering a structured and logical approach to comprehending complex image compositions.
\section{Experiments}
\label{sec:experiment}

\begin{table*}[t!]
\caption{PartImageNet \textit{val} set and Pascal-Part-108 \textit{test} set results. mIoU on parts and super-category, mAvg are reported. Reported results are averaged over 3 runs.
}
\centering
\subfloat[
\textbf{PartImageNet \textit{val} set results}
\label{tab:partimagenet}
]{
\centering
\begin{minipage}{0.59\linewidth}{\begin{center}
\tablestyle{4pt}{1.2}
\scalebox{0.8}{
\begin{tabular}{l|c|cc}
\multirow{2}{*}{method} & \multirow{2}{*}{backbone} & \multicolumn{2}{c}{mIoU} \\ \cline{3-4} 
 &  & \multicolumn{1}{c|}{Part} & Super-Category \\ \hline
DeepLabv3+~\cite{chen2018encoder} & ResNet-50~\cite{he2016deep} & \multicolumn{1}{c|}{60.57} & - \\
MaskFormer~\cite{cheng2021per} & ResNet-50~\cite{he2016deep} & \multicolumn{1}{c|}{60.34} & - \\
Compositor~\cite{he2023compositor} & ResNet-50~\cite{he2016deep} & \multicolumn{1}{c|}{61.44} & - \\
kMaX-DeepLab~\cite{yu2022k} & ResNet-50~\cite{he2016deep} & \multicolumn{1}{c|}{65.75} & 89.16 \\ 
\modelname & ResNet-50~\cite{he2016deep} & \multicolumn{1}{c|}{\textbf{67.83}} & \textbf{90.41} \\ \hline
SegFormer~\cite{xie2021segformer} & MiT-B2~\cite{xie2021segformer} & \multicolumn{1}{c|}{61.97} & - \\
MaskFormer~\cite{cheng2021per} & Swin-T~\cite{liu2021swin} & \multicolumn{1}{c|}{63.96} & - \\
Compositor~\cite{he2023compositor} & Swin-T~\cite{liu2021swin} & \multicolumn{1}{c|}{64.64} & - \\
kMaX-DeepLab~\cite{yu2022k} & ConvNeXt-T~\cite{liu2022convnet} & \multicolumn{1}{c|}{68.52} & 91.34 \\
\modelname & ConvNeXt-T~\cite{liu2022convnet} & \multicolumn{1}{c|}{\textbf{70.31}} & \textbf{92.65} \\
\end{tabular}
}
\end{center}}\end{minipage}
}
\subfloat[
\textbf{Pascal-Part-108 \textit{test} set results}
\label{tab:pascal}
]{
\centering
\begin{minipage}{0.39\linewidth}{\begin{center}
\tablestyle{4pt}{1.2}
\scalebox{0.8}{
\begin{tabular}{l|c|c}
method & Part mIoU & mAvg \\ \hline
SegNet~\cite{badrinarayanan2017segnet} & 18.6 & 20.8 \\
FCN~\cite{long2015fully} & 31.6 & 33.8 \\
DeepLab~\cite{chen2017deeplab} & 31.6 & 40.8 \\
DeepLabv3+~\cite{chen2018encoder} & 46.5 & 48.9 \\
BSANet~\cite{zhao2019multi} & 42.9 & 46.3 \\
GMNet~\cite{michieli2020gmnet} & 45.8 & 50.5 \\
FLOAT~\cite{singh2022float} & 48.0 & \textbf{53.0} \\ 
HSSN~\cite{li2022deep} & 48.3 & - \\
DeepLabv3+~\cite{chen2018encoder}+ LOGICSEG~\cite{li2023logicseg} & 49.1 & - \\ \hline 
kMaX-DeepLab~\cite{yu2022k} & 48.3 & 49.9 \\
\modelname & \textbf{49.8} & 52.0
\end{tabular}
}
\end{center}}\end{minipage}
}
\end{table*}

In this section, we first provide the experimental setup, followed by the main results on PartImageNet~\cite{he2022partimagenet} and Pascal-Part-108~\cite{michieli2020gmnet}. We conduct ablation studies on PartImageNet to demonstrate the effectiveness of our designs. We also provide visualizations to better understand \modelname.

\subsection{Experimental Setup}
\textbf{Datasets}\quad We conduct experiments on two popular object parsing benchmarks: PartImageNet~\cite{he2022partimagenet} and PASCAL-Part-108~\cite{michieli2020gmnet}. We provide the detailed statistics of each dataset and the class definitions below:
\begin{itemize}
    \item PartImageNet~\cite{he2022partimagenet} contains 24095 elaborately annotated general images from ImageNet~\cite{deng2009imagenet}, which are split into 20481/1206/2408 for \textit{train}/\textit{val}/\textit{test}. It is associated with 40 part classes, which are grouped into 11 object classes following the official class definition.
    \item Pascal-Part-108~\cite{michieli2020gmnet} expands upon the part definition introduced in Pascal-Part-58~\cite{chen2014detect}, providing a more intricate benchmark with finer part-level details. This extension maintains the original split of VOC~\cite{pascal-voc-2010} and encompasses a dataset of 10,103 images across 20 object classes and 108 part classes. Our experiments adhere to the original split, utilizing 4,998 images for training and 5,105 images for testing.
\end{itemize}

\noindent \textbf{Evaluation Metrics}\quad
We evaluate the performance of \modelname on the PartImageNet dataset~\cite{he2022partimagenet} using the mean Intersection over Union (mIoU) on both part and super-category levels. It's important to note that for PartImageNet, we choose to report performance on the super-category level because the parts in PartImageNet are defined within the context of super-category. The hierarchy of super-category is inherited for training \modelname on this dataset.
In the case of Pascal-Part-108, our evaluation includes reporting part mIoU, and additionally, we calculate the mAvg on the object level. The mAvg metric, as defined in the literature~\cite{zhao2019multi}, provides the average mIoU score of all parts belonging to an object. We refer the reader to FLOAT~\cite{singh2022float} for a detailed explanation of these metrics.

\noindent \textbf{Training details}\quad 
We implement \modelname based on the kMaX-DeepLab architecture~\cite{yu2022k}, utilizing its official PyTorch re-implementation codebase. To ensure a fair comparison, we adopt the training settings from kMaX-DeepLab. 
The backbone, pretrained on ImageNet~\cite{he2016deep,liu2022convnet}, followed a learning rate multiplier of 0.1. For regularization and augmentations, we incorporate drop path~\cite{huang2016deep} and random color jittering~\cite{cubuk2018autoaugment}. The optimizer used is AdamW~\cite{kingma2014adam,loshchilov2018decoupled} with a weight decay of 0.05.
Unless otherwise specified, we train all models with a batch size of 64 on a single A100 GPU, performing 40,000 iterations on PartImageNet~\cite{he2022partimagenet} and 10,000 iterations on Pascal-Part-108~\cite{chen2014detect}. The first 2,000 and 500 steps serve as the warm-up stage, where the learning rate linearly increases from 0 to $5 \times 10^{-4}$.
The training objective for \modelname includes the combination of kMaX-DeepLab's original losses and the proposed contrastive loss terms, as specified in Eq.~\ref{formula:contrast_p}, Eq.~\ref{formula:contrast_o} and Eq.~\ref{formula:contrast_cross}:
\begin{align*}
\mathcal{L} = &\lambda_{\text{kMaX}}\mathcal{L}_{\text{kMaX}} + \lambda_{p\_con}\mathcal{L}_{p\_con} + \\
&\lambda_{o\_con}\mathcal{L}_{o\_con} + \lambda_{logic}\mathcal{L}_{logic}.
\end{align*}
Here, $\mathcal{L}_{\text{kMaX}}$ represents the loss from kMaX-DeepLab~\cite{yu2022k}, and $\lambda_{\text{kMaX}}$ follows the official setting. The weights for the proposed loss terms are set to $\lambda_{p\_con} = 2$, $\lambda_{o\_con} = 2$, and $\lambda_{logic} = 1$.
\modelname uses the exact same number of part and object queries corresponding to the part and object classes in the dataset. Specifically, we set $P$ to 41 and 109, and $\widetilde{P}$ to 12 and 21 (with one additional learnable component for representing the background at both the part and object levels) in PartImageNet and Pascal-Part-108, respectively. This design enables a straightforward and highly interpretable inference process, using nearest neighbor search for parts and objects separately during inference. Afterward, we compute the top-scoring logical path and reassign the predicted classes based on that path.

\subsection{Main Results}
Our main results on the PartImageNet~\cite{he2022partimagenet} \textit{val} set and PASCAL-Person-Part~\cite{xia2017joint} \textit{test} set are summarized in Tab.~\ref{tab:partimagenet} and Tab.~\ref{tab:pascal}, respectively.

\noindent \textbf{PartImageNet \textit{val} set}\quad In Table~\ref{tab:partimagenet}, we present a comparison between \modelname and several classic segmentation models on the PartImageNet \textit{val} set. As a strong cluster-based mask transformer, kMaX-DeepLab~\cite{yu2022k} already surpasses previous works by a substantial margin. Particularly, with a ResNet-50~\cite{he2016deep} as the backbone, \modelname achieves a significant 2.08\% improvement in part mIoU over kMaX-DeepLab. Using the more powerful ConvNeXt-Tiny~\cite{liu2022convnet} as the backbone, \modelname further elevates the performance to 70.31 part mIoU, surpassing kMaX-DeepLab~\cite{yu2022k} with the same backbone by 1.79\% part mIoU. 
Notably, \modelname consistently enhances super-category segmentation results in comparison to kMaX-DeepLab. With ResNet-50 as the backbone, we observe an improvement of 1.25\%, while with ConvNeXt-Tiny, the enhancement reaches 1.31\%.

\noindent \textbf{Pascal-Part-108 \textit{test} set}\quad In Tab.~\ref{tab:pascal}, we summarize \modelname's performance on Pascal-Part-108 \textit{test} set against other methods. All the models utilize ResNet-101~\cite{he2016deep} as the backbone. As observed in the table, \modelname achieves the best performance, setting new state-of-the-art results with 49.8 part mIoU. Notably, \modelname outperforms the previous state-of-the-art method LOGISEG~\cite{li2023logicseg} and the strong baseline kMaX-DeepLab~\cite{yu2022k} by a substantial 0.7\% and 1.5\% in part mIoU, respectively. In terms of object segmentation, \modelname demonstrates a notable improvement over kMaX-DeepLab, achieving a substantial 2.1\% increase. This underscores \modelname's capability not only to refine parsing to a finer granularity but also to enhance overall segmentation quality.

\begin{figure*}[t!]
    \centering
    \includegraphics[width=\textwidth]{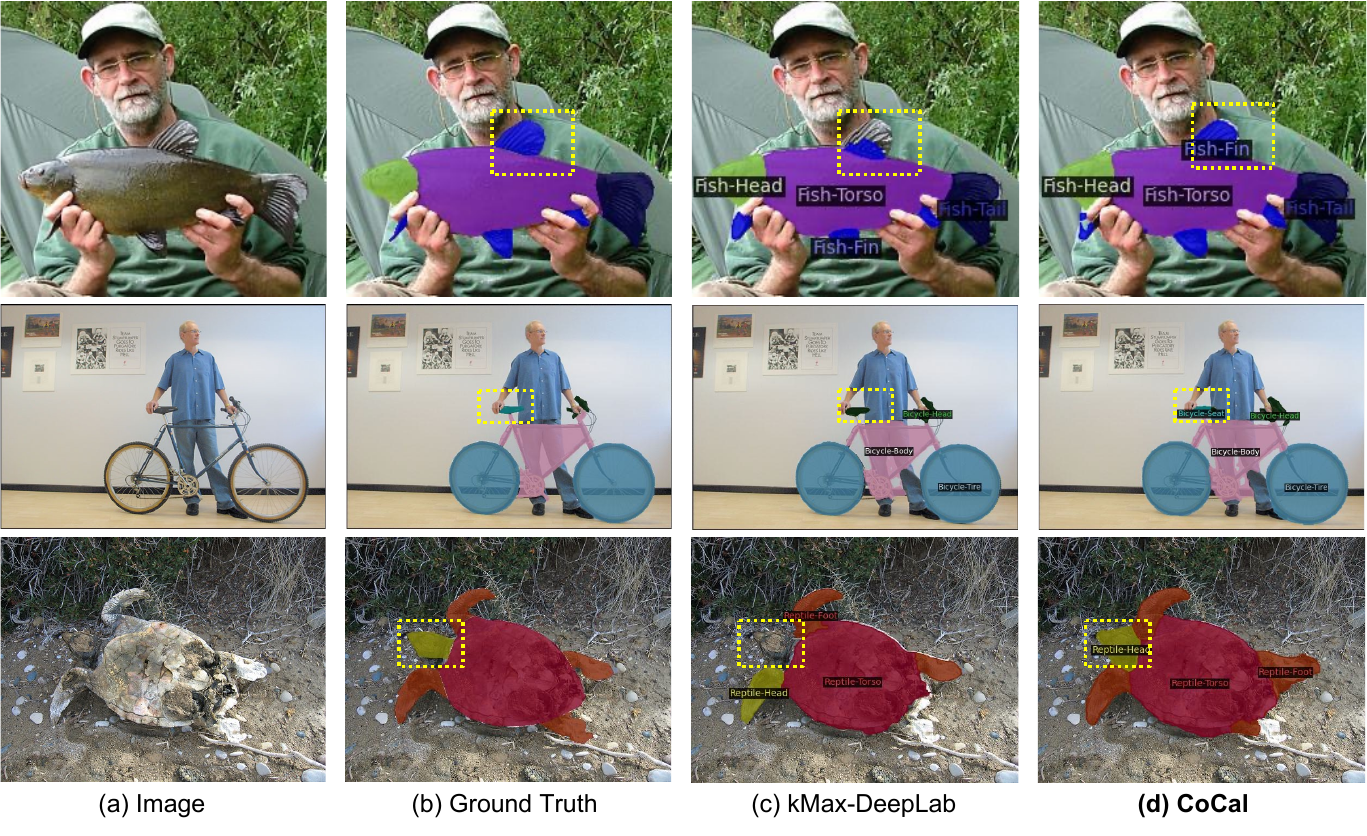}
    \caption{\textbf{Qualitative comparison for \modelname and kMaX-DeepLab on PartImageNet.} Note that \modelname produces much more accurate object parsing results with precise boundaries (\eg, row 1) and fewer missed detections (\eg, row 2 \& 3).}
    \label{fig:vis}
\end{figure*}

\subsection{Qualitative Results}
Fig.~\ref{fig:vis} depicts three representative visual results on PartImageNet. As seen, \modelname yields better object parsing results compared to kMaX-DeepLab~\cite{yu2022k} by yielding more accurate boundaries (\eg, row 1) and detecting parts that are missed by kMaX-DeepLab (\eg, row 2 \& 3). 

\subsection{Ablation Studies}
\label{subsec:ablation-studies}

\begin{table}[t]
% \small
\centering
\caption{Ablation study on individual module designs for \modelname on PartImageNet \textit{val} set. All models use ResNet-50~\cite{he2016deep}.}
\tablestyle{3pt}{1.0}
\begin{tabular}{l|cccc|c}
method & Dictionary & $\mathcal{L}_{p\_con}$ & $\mathcal{L}_{o\_con}$ & $\mathcal{L}_{logic}$ & Part mIoU \\ \hline
 & \XSolidBrush & \XSolidBrush & \XSolidBrush & \XSolidBrush & 65.75 \\
 & \checkmark & \XSolidBrush & \XSolidBrush & \XSolidBrush & 64.31 \\
 \modelname & \checkmark & \checkmark & \XSolidBrush & \XSolidBrush & 65.87 \\ 
 & \checkmark & \checkmark & \checkmark & \XSolidBrush & 66.53 \\
 & \checkmark & \checkmark & \checkmark & \checkmark & \textbf{67.83}    
\end{tabular}
\label{tab:moduleablate}
\end{table}

\noindent \textbf{Evaluating the Impact of Dictionary Components, Contrastive Components, and Logical Constraints}\quad In Table~\ref{tab:moduleablate}, we conduct ablation studies to assess the impact of our core design components on \modelname. Our findings reveal that simply adapting kMaX-DeepLab to our proposed dictionary-based mask transformer, by incorporating dictionary components, does not inherently enhance performance. In fact, the model's part mIoU on PartImageNet declines from 65.75 to 64.31. This drop is attributed to the insufficient discriminative power of the dictionary components, which, in the absence of contrastive loss supervision, leads to ambiguity during the nearest neighbor search among similar parts.
We then incorporate contrastive learning objectives into the dictionary-based mask transformer, resulting in a marked improvement of 2.22\% in part mIoU. Specifically, adding contrastive supervision on parts through $\mathcal{L}_{p\_con}$ brings a 1.56\% improvement, while additional contrast on object-level targets $\mathcal{L}_{o\_con}$ brings another 0.66\% improvement. Notably, this performance surpasses the baseline kMaX-DeepLab by 0.78\% in mIoU, supporting our hypothesis that cultivating a discriminative dictionary is crucial for the effective functioning of the dictionary-based mask transformer.
In the final phase of our ablation study, we integrate logical constraints into the model, which brings a notable 1.30\% improvement, establishing a new state-of-the-art performance on the PartImageNet \textit{val} set with ResNet-50.

\noindent \textbf{Impact of Memory Bank Size $S$}\quad Table~\ref{tab:memory} examines the effect of varying the size of the memory bank. A notable observation is the performance degradation when $S$ is reduced to 50. This decline suggests that a smaller memory bank size is inadequate in providing a sufficient number of samples for effective contrastive learning objectives. Conversely, expanding the memory bank size to 150 and 200 also results in a gradual decrease in performance. This decline could be attributed to the limited diversity of instances in the dataset. In such cases, an oversized memory bank may lead to redundancy in samples, which adversely affects the learning process for the dictionary components. This finding underscores the importance of optimizing the memory bank size to balance the need for sufficient sample diversity without introducing detrimental redundancy.

\noindent \textbf{Influence of the Number of Negative Samples $k$}\quad In Table~\ref{tab:negative}, we examine the influence of varying the number of negative samples, denoted as $k$. The findings illustrate a discernible trend: an insufficient number of negative samples corresponds to a decline in performance from 67.83 to 66.28 part mIoU. This suggests that a limited pool of negative samples may not provide sufficient challenge or diversity to effectively train the model. Conversely, excessively increasing the number of negative samples can also have detrimental effects. Specifically, an overabundance of negatives can lead to a scenario where the model's learning is dominated by 'easy' negatives, ultimately resulting in suboptimal performance.

\begin{table}[t!]
\caption{Ablation study on number of memory bank size $S$ and negative samples $k$ for \modelname with ResNet-50 as backbone on PartImageNet \textit{val} set.}
\centering
\subfloat[
\textbf{Ablation on number of memory bank size $S$}
\label{tab:memory}
]{
\centering
\begin{minipage}{0.49\linewidth}{\begin{center}
\tablestyle{3pt}{1.2}
\scalebox{1.0}{
\begin{tabular}{c|c}
    \# memory bank $S$ & Part mIoU \\ \hline
    50                 & 66.50 \\
    100                & \textbf{67.83} \\
    150                & 67.16 \\
    200                & 67.02  
\end{tabular}
}
\end{center}}\end{minipage}
}
\subfloat[
\textbf{Ablation on number of negative samples $k$}
\label{tab:negative}
]{
\centering
\begin{minipage}{0.49\linewidth}{\begin{center}
\tablestyle{3pt}{1.2}
\scalebox{1.0}{
\begin{tabular}{c|c}
    \# negative sample $k$ & Part mIoU \\ \hline
    50                & 66.28 \\
    100               & \textbf{67.83} \\
    200               & 66.40 \\
    all               & 65.74 
\end{tabular}
}
\end{center}}\end{minipage}
}
\end{table}

\begin{table}[h]
% \vspace{-3mm}
% \scriptsize
\centering
\caption{Performance of \modelname with different baselines using ResNet-50 as backbone on PartImageNet \textit{val} set.}
\begin{tabular}{l|cc}
\multirow{2}{*}{method} & \multicolumn{2}{c}{mIoU} \\ \cline{2-3} 
 & \multicolumn{1}{c|}{Part} & Super-Category \\ \hline
MaskFormer~\cite{cheng2021per} & \multicolumn{1}{c|}{60.34} & - \\
CoCal (MaskFormer) & \multicolumn{1}{c|}{63.52} & 86.67 \\ \hline
Mask2Former~\cite{cheng2022masked} & \multicolumn{1}{c|}{63.62} & 87.20 \\
CoCal (Mask2Former) & \multicolumn{1}{c|}{66.39} & 88.72
\end{tabular}
% \vspace{-6mm}
\label{tab:baseline}
\end{table}

\noindent \textbf{Generalizability of \modelname}\quad In Table~\ref{tab:baseline}, we evaluate the generalizability of \modelname using two baseline models. Incorporating \modelname into MaskFormer~\cite{cheng2021per} and Mask2Former~\cite{cheng2022masked} results in part mIoU improvements of 3.18 and 2.77, respectively. To be more specific, we change the cross-attention in MaskFormer to soft clustering attention~\cite{he2023compositor} in order to integrate with \modelname. Besides, we decrease the number of queries by changing Hungarian Matching to fixed matching mechanism. This experiment illustrates that \modelname can seamlessly integrate into various modern segmentation frameworks, consistently enhancing performance across different architectures.

\section{Conclusion}
In conclusion, this paper introduces \modelname, an innovative model for object parsing that is rooted in a dictionary-based framework. A key aspect of \modelname is its emphasis on elucidating the intrinsic relationships between parts and objects, which significantly enhances the interpretability and consistency of parsing outcomes. Building upon an off-the-shelf cluster-based mask transformer, \modelname introduces the dictionary-based mask transformer by incorporating dictionary components. These components are associated with their corresponding classes in a fixed one-to-one manner. By implementing a component-wise contrastive algorithm and logical relation modeling, \modelname aligns its parsing predictions more closely with the underlying semantic hierarchy, akin to human cognitive processing. The consistency in prediction is further enhanced by the proposed post-processing function. This approach not only improves the accuracy of the parsing but also provides a deeper understanding of the complex interplay between part and object entities in images. As a result, \modelname sets the new state-of-the-art performances on PartImageNet and Pascal-Part-108 and surpasses prior arts by a non-trivial margin.

\newpage
{
    \small
    \bibliographystyle{ieeenat_fullname}
    \bibliography{main}
}


\end{document}